# Prediction Model for Mortality Analysis of Pregnant Women Affected With COVID-19

Quazi Adibur Rahman Adib
*Department of Computer Science and Engineering*
*Brac University*
Dhaka, Bangladesh
E-mail: quazi.adibur.rahman.adib@g.bracu.ac.bd

Sidratul Tanzila Tasmi
*Department of Computer Science and Engineering*
*Islamic University of Technology*
Gazipur, Bangladesh
E-mail: sidratultanzila@iut-dhaka.edu

Md. Shahriar Islam Bhuiyan
*Department of Computer Science and Engineering*
*Islamic University of Technology*
Gazipur, Bangladesh
E-mail: shojeb.bhuiyan@gmail.com

Md. Mohsin Sarker Raihan
*Department of Biomedical Engineering*
*Khulna University of Engineering and Technology*
Khulna, Bangladesh
E-mail: msr.raihan@gmail.com

Abdullah Bin Shams
*Department of Electrical and Computer Engineering*
*University of Toronto*
Toronto, Canada
E-mail: abdullahbinshams@gmail.com

*Abstract*—COVID-19 pandemic is an ongoing global pandemic which has caused unprecedented disruptions in the public health sector and global economy. The virus, SARS-CoV-2 is responsible for the rapid transmission of coronavirus disease. Due to its contagious nature, the virus can easily infect an unprotected and exposed individual from mild to severe symptoms. The study of the virus's effects on pregnant mothers and neonatal is now a concerning issue globally among civilians and public health workers considering how the virus will affect the mother and the neonate's health. This paper aims to develop a predictive model to estimate the possibility of death for a COVID-diagnosed mother based on documented symptoms: dyspnea, cough, rhinorrhea, arthralgia, and the diagnosis of pneumonia. The machine learning models that have been used in our study are support vector machine, decision tree, random forest, gradient boosting, and artificial neural network. The models have provided impressive results and can accurately predict the mortality of pregnant mothers with a given input.The precision rate for 3 models(ANN, Gradient Boost, Random Forest) is 100% The highest accuracy score(Gradient Boosting,ANN) is 95%,highest recall(Support Vector Machine) is 92.75% and highest f1 score(Gradient Boosting,ANN) is 94.66%. Due to the accuracy of the model, pregnant mother can expect immediate medical treatment based on their possibility of death due to the virus. The model can be utilized by health workers globally to list down emergency patients, which can ultimately reduce the death rate of COVID-19 diagnosed pregnant mothers.

*Index Terms*—Machine Learning, Mortality Analysis, Pregnant Women, COVID-19, Random Forest, Support Vector Machine, Artificial Neural Network, Gradient Boosting, Decision Tree, SMOTE

## I. INTRODUCTION

Coronavirus (COVID-19) is a viral and infectious disease caused due to SARS COV2 virus.The disease has been rapidly spreading across the world causing a severe public health crisis. The precise extension of the risk of this disease is still under research. The disease has infected people of all ages and has caused a catastrophe on the economic and health sectors. As of 18 November 2021, there have been 254,847,065 confirmed cases and 5,120,712 deaths [1]. The disease can be transmitted to anyone through close contact with affected patients via aerosol and respiratory droplets of the infected person [2]. People who have medical complications such as diabetes, cancer and chain smoking have more critical condition than rest of the mass [3]. Due to the contagious nature of the spreading of virus through air, it is recommended to avoid mass gathering and maintain social distancing in order to avoid being infected from the virus. However, people from countless life sectors including emergency health workers constantly have to stay on duty and social isolation under these circumstances is almost impossible. Men, women, especially pregnant women who are working outside are more likely to be exposed to the virus. The complication and the extent to which the virus affects pregnant mothers is still a big concern to not just for the mother but the neonatal. The pandemic has caused disruption to the mental health of people all over and for pregnant women, the anticipation and excitement of the neonate have been replaced with stress and anxiety following up the infection of coronavirus. A model for early prediction of the likelihood of being contact with COVID-19 is essential in order to take necessary precautions for early recovery and minimize the spread [4]. However, due to the increasing fatality rate of COVID-19 affected people around the world, it is essential to note down emergency patients based on the symptoms of the disease. Therefore, the study of the outcome of COVID-19 on pregnant women is a growing need for both the mother and the newborn. A painless and risk-free methodology for detection of COVID-19 contact in pregnant mother can lower the health risk for both the mother and neonate [5]. Pregnant women with a mild cold or flu-like symptoms can recover properly by maintaining social isolation, medication, and care. However, women with moderate to severe symptoms require to be hospitalized. The duration in which trimester the virus has affected the mother is a critical factor in determining the risk of the mother's body. Some of the severe symptoms on an affected mother include odynophagia, chills, severe cough, and arthralgia [6]. The deleterious effects of COVID-19 can lead to the potential death of the mother. A predictive analysis of the potential life-threatening cases for individual patients can help doctors make critical clinical decisions to reduce the overall mortality rate and minimize the patients' health impairment. The primary focus of our study is to build a predictive model for individual patient's life-threatening risk and mortality based on severe symptoms. The machine can accurately predict the patient's fatal mortality. With the wide availability and cost-effectiveness of this model in the public health sector, doctors can categorize patients based on their severity and assist with proper medication. The limited ICU facilities and the crisis of health service to affected patients can be managed and patients with high life threatening risks and probability can be saved. This can overall reduce the mortality rate and the damage caused due to corona virus can be cut down.

## II. Related Works

An article regarding the threatning state of pregnant mother affected with COVID-19 was first published on April 22,2021 [7]. The article gives an idea of the severity of the virus on pregnant mother. It concludes that pregnant mother are 20 times more at risk of fatal death cases during pregnancy than normal healthy mother.

A paper regarding the death cases due to COVID-19 GPR and ANN-based prediction models for COVID-19 death cases was published in the 2020 International Conference on Communications, Computing, Cybersecurity, and Informatics [3]. The paper aimed to build a model to predict the number of estimated affected cases based on the the data of people with following criteria: older people, smoking, with Process Regression (GPR) as well as Artificial Neural Network Model(ANN). The data set for this model has been collected from the daily update of estimated death cases from WHO and the data has been modeled towards older people, smoke and diabetic concerning patients. The paper proposes the implementation of GPR and ANN models to estimate the mortality rate, which can also make a correlation among old age, diabetes, and smoking with the disease and the cause of death. The model can also give an overall view of the number of possible cases of COVID in the future and the duration of the pandemic to last.

The outcomes of COVID-19 on pregnant women have also been studied in the paper 'Maternal and Neonatal Morbidity and Mortality Among Pregnant Women With and Without COVID-19 Infection, The INTERCOVID Multinational Cohort Study' published on January 21, 2021 [8]. The main objective of the study was to learn about the health outcomes of both healthy and COVID-19 affected mothers. A consistent association exists between pregnant individuals with COVID-19 diagnosis and higher rates of adverse outcomes, including maternal mortality, preeclampsia, and preterm birth compared with pregnant individuals without COVID-19 diagnosis. The result of the study stated that women with COVID-19 have a higher cesarean delivery rate but lower rate of spontaneous initiation of labor, resulting in complications of COVID affected mothers during delivery. The study states that having dyspnea, chest pain, and cough with fever increased the risk for severe maternal health and on neonatal's birth. Women with obesity have a higher risk of complications.

In the proceeding paper, 'Analysis of pregnancy outcomes in pregnant women with COVID-19 in Hubei Province' published in March 2020 at Chinese Journal of Obstetrics and Gynecology [9], the effects of COVID-19 on the pregnant woman and neonatal prognosis was studied. The study was conducted on 16 pregnant women with 1 severely COVID- affected mother. The study concluded that along with the timely termination of pregnancy, the risk of premature birth of the newborn will not be affected. Furthermore, it implied that treatment is essential to rehabilitate the newborn mother. We can conclude this study that the newborn can have no risk from the COVID-19 affected mother with timely treatment. The proper diagnosis and timely medication are however necessary.

A study regarding the outcomes of giving birth while being affected with COVID-19 was studied in the paper 'Characteristics and Outcomes of Women With COVID-19 Giving Birth at US Academic Centers During the COVID-19 Pandemic' [10]. The study included certain pregnancy related parameters to have an overall statistical analysis.The study concluded that pregnant woman affected with COVID-19 are more likely to deliver baby than healthy woman by 37 weeks. Additionally, the woman affected with COVID-19 had a significantly higher mortality rate. A study was conducted regarding the overview of the maternal deaths in Brazil, 'Maternal Mortality and COVID-19' published on 16 July,2021 [11]. The data set was collected from the Brazilian Ministry of health's surveillance system. The dataset included: timing of symptom onset and death (pregnancy or postpartum), gestational age, mode of birth, maternal age, comorbidities, and/or risk factors. The study concluded that the external environmental situation in Brazil can be a variable factor in causing maternal deaths, in comparison to other parts of the world. Antenatal care and public health management systems can also factor in the mortality rate of pregnant women from country to country. Thus, proper categorization of mothers based on their severity of health complications is essential.

Finally, a study conducted regarding the clinical complications and death among pregnant women with COVID-19

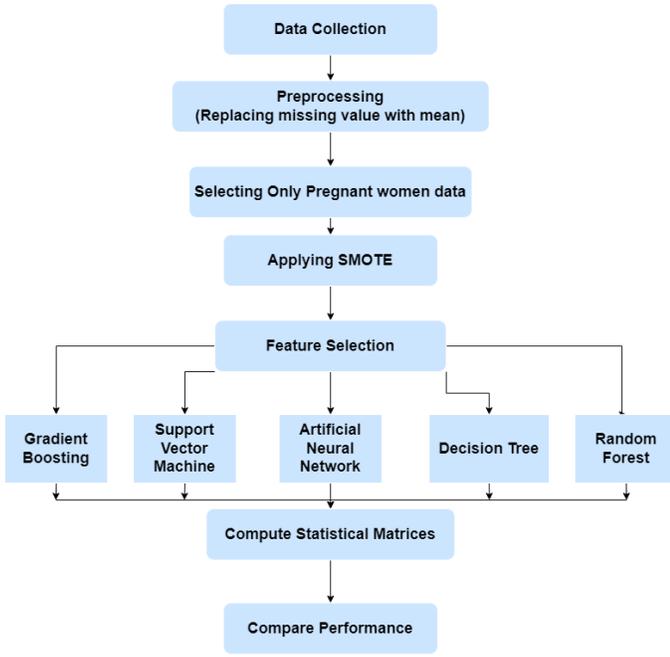

Fig. 1: Workflow of this work

was analyzed in the paper 'The risk of clinical complications and death among pregnant women with COVID-19 in the Cerner COVID-19 cohort: a retrospective analysis, published on 16 April 2021 [12]. The data set was collected from Cerner COVID-19 De-Identified Data Cohort. The study focused on the extent at which the virus deteriorates the mothers' health and outcomes touch as the maximum length of hospital stay, moderate ventilation, invasive ventilation, and death.It is observed that mothers affected with COVID-19 are more likely to be hospitalized than healthy mothers COVID-19 affected mothers are highly in need of mechanical ventilator receipts and medical care. The study concluded that with sufficient care and treatment, COVID-19 diagnosed pregnant mothers can have very few complications and minimize overall mortality rate.

## III. METHODOLOGY

In this research, a wide range of machine learning classifiers has been used for prediction. Besides, the data-set required pre-processing, feature selection, SMOTE, algorithms.

### A. Data Collection

The documented data set has been collected from the article 'A Machine Learning Approach as an Aid for Early COVID-19 Detection' [6], published on 18 June 2021. The data set is up to date with the recent year. The documented data has been processed with a series of steps. We have primarily focused and collected the data associated with pregnant women specifically from all the categories of the following data set for our research.

### B. Data Preprocessing

Data pre-processing was a crucial step for this work. The collected data set has been cut short to only a specific columns were selected. A number of steps were followed to increase the precision rate of the model from the following data set columns. From the following pieces of information, the null values for some rows were modified and replaced by the mean values of that particular column.

### C. SMOTE

SMOTE stands for Synthetic Minority Oversampling Technique. It is an oversampling technique where the main goal is to generate minority class data. Initially, the algorithm sets the number of observations and select class distribution. After that, it generates data randomly using KNN [13] to choose synthetic data.

Machine Learning models do not work well when there are imbalances in the training dataset. Augmenting new data using various techniques solves these imbalances. In our work, we used SMOTE [14] to augment data. After applying SMOTE, both classes' ratio was 50:50.

### D. Data Splitting

We have split our data into two sets. One is for training and another is for testing. For training we have used 70% data and for testing we have used 30%.

### E. Feature Selection

Our paper emphasized on the following symptoms of the affected mother: COVID, Odynophagia, Chills, Arthralgia, Rhinorrhea, Pneumonia, Cough, Dyspnea and type of patient (Ambulatory/ Hospitalized).

### F. Algorithms

*1) Support Vector Machine:* Support Vector Machine [15] is a classical machine learning algorithm. The central idea of Support Vector Machine is to generate higher dimensional hyper-plane by applying statistical frameworks to classify different classes. Here, the algorithm tries to maximize the distance between different classes in higher dimensional hyper-plane. By doing this model can ensure its accuracy.

*2) Decision Tree:* Decision Tree [16] is a machine learning algorithm that is both powerful and simple. It forms a tree-like structure in which the branches represent the test results and the nodes denote any decisions that need to be made. The initial training set is evaluated and partitioned into smaller groups based on defined parameters.This procedure is referred to as 'Recursive Partitioning' as it is done repeatedly. The prediction is reached by recursively using the basic induction method. The input samples are searched in a 'top-down' manner to test every attribute at each node. Afterwards, entropy and information gain are calculated to identify which attribute to test at each node. This information is then used to make prediction.

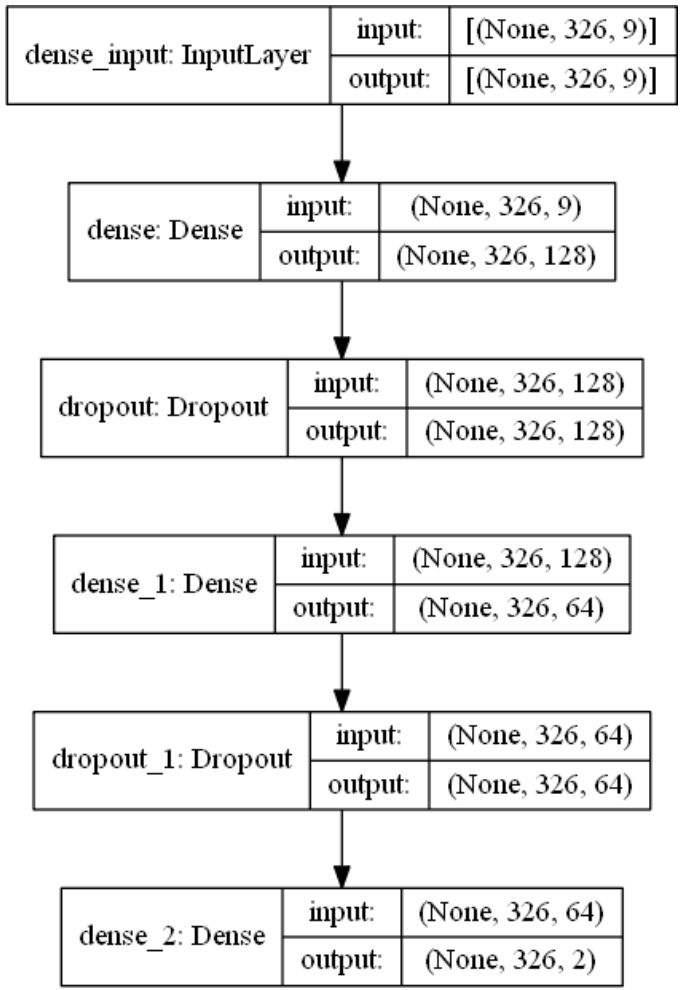

Fig. 2: ANN Architecture

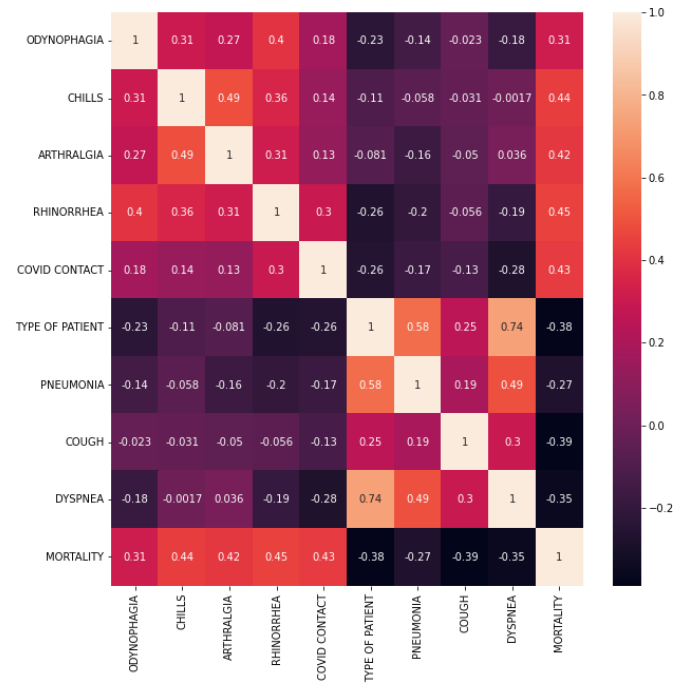

Fig. 3: Correlation Matrix

*3) Random Forest:* Random Forest [17] is an ensemble learning model. Random Forest combines bootsrap aggregation and random feature selection to construct a set of decision trees. The newly created trees are used for the training procedure. The predicted class of the majority of the trees is selected as the output of the random forest. Random Forest can make quick and accurate predictions on very large datasets.

*4) Gradient Boosting:* Gradient Boosting is a popular ensemble machine learning technique. To execute prediction tasks, Gradient Boosting uses decision trees [18]. This model creates several trees one by one, each one attempting to improve on the preceding one. As a result, it becomes a boosting algorithm. Gradient boosting optimization technique is used to optimize this model.

*5) Artificial Neural Network:* Artificial Neural Networks (ANNs) [19] are often referred as Neural Networks. ANN's architecture is inspired by a biological brain. It is a structured like a multi-layered weighted directed graph. It is made up of nodes, edges and non-linear activation functions.

We employed a Multi-layer Artificial Neural Network. At first, we had an input layer that links to a fully connected dense layer [20]. That layer's unit is 128. In addition, we used the ReLU [21] activation function as a non-linear activation function. To minimize overfitting, this layer was connected to a dropout layer [22]. After that, the outputs of previous layer got into another dense layer followed by another dropout layer. Second dense layer's unit is 64. Finally, an output layer with Softmax non-linear activation function was implemented [23]. We utilized Adam optimizer [24] and Sparse Categorical Cross Entropy [25] as a loss function to construct the model.

IV. RESULTS AND DISCUSSION

The correlation matrix aids in visualizing the interrelationships between the data-set's parameters. The correlation of the parameters are expressed within the range -1 to +1. Values within +0.1 to +1 denote a strong positive correlation where values within -0.1 to -1 denote a strong negative correlation. Values closer to 0 indicates insignificant correlation between the parameters.

According to the correlation matrix in Fig. 3, the correlation between mortality and COVID contact is 0.43, which indicates a significantly strong positive correlation. This indicates that the presence of the virus in pregnant mothers further risk their life. Rhinorrhea, a disease caused due to inflammation in the nasal tissue, has the highest correlation with mortality, expressed by the value 0.45. Chills also have a very high correlation value of 0.44 with mortality. Chills and Rhinorrhea also have a strong positive correlation with COVID contact, with respective values of 0.14 and 0.3. The presence of Rhinorrhea and chills is an identifier for the presence of COVID-19 which is deeply correlated with mortality. This can be further emphasized from the paper 'Maternal and Neonatal

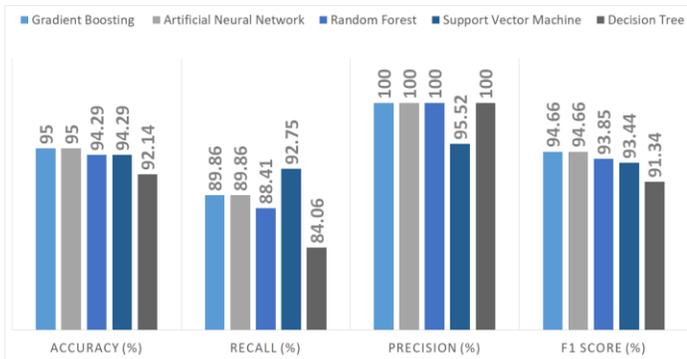

Fig. 4: Model Performances

Morbidity and Mortality Among Pregnant Women With and Without COVID-19 Infection, The INTERCOVID Multinational Cohort Study' [8], which concludes the symptoms of cough, chest pain, dyspnea increases the fatal condition for pregnant mothers. Odynophagia, a disease related to the pain in swallowing and morality has a correlation value of 0.31, which denotes a very strong positive correlation. Arthralgia which causes pain in joints, and mortality have a correlation value of 0.42. These symptoms further emphasize the strong positive correlation between COVID and mortality. The major symptoms admissible to a COVID-19 affected mother was chosen to build the model. However, the background, race and previous clinical issues of the pregnant mother: obesity, diabetes, smoking have not been included in our data set.

Fig. 4, illustrates the results and performance of this study. Support Vector Classifier had an accuracy of 94.29%, precision of 95.52%, recall of 92.75% and F1 score of 94.12%. Decision Tree had an accuracy of 92.14%, precision of 100%, recall of 84.06% and F1 score of 91.34%. Random Forest performed at an accuracy of 94.29%, precision of 100%, recall of 88.41% and F1 score of 93.85%. Gradient Boosting had a prediction accuracy of 95%, precision of 100%, recall of 89.86% and F1 score of 94.66%. Our ANN model performed at an accuracy of 95%, precision of 100%, recall of 89.86% and F1 score of 94.66%.

In our study, the most important score to consider is the precision score as we are more concerned about being precise about predicting the mortality. So, Decision Tree, Random Forest, Gradient Boosting and ANN are ideal models as they all have a perfect precision score of 100%. Further based on F1 score we can limit our choice of models to Gradient Boosting and ANN.

## V. CONCLUSION

In this study, our main objective was to build a machine learning based predictive model for mortality prediction of pregnant women affected with COVID. Here, Decision Tree, Random Forest, Gradient Boosting and ANN proved to be the ideal models to be the most precise in predicting mortality. But considering all the important metrics in this study, Gradient Boosting and ANN are the best performing models. Moreover, we have also looked at the relation between the mortality of pregnant women affected with COVID and other factors such as Odynophagia, Chills, Arthralgia, Rhinorrhea, Pneumonia, Cough, Dyspnea and whether the patient was ambulatory or hospitalized. The model can be accessed by any health worker and based on the predictions the health care workers can take better precautions and intensive care for the pregnant women. This will result to better distribution of limited ICU beds and mechanical ventilation which can contribute to the overall decrease of the total mortality rate of pregnant women affected with COVID.